\pdfoutput=1

\documentclass[11pt]{article}

\usepackage[]{acl}

\usepackage{times}
\usepackage{latexsym}

\usepackage{times}
\usepackage{latexsym}
\usepackage{url}
\usepackage{amsfonts}
\usepackage{multirow}
\usepackage{amssymb}
\usepackage{subfigure}
\usepackage{graphicx}
\usepackage{arydshln}
\usepackage{amsmath}
\usepackage{caption}
\usepackage{CJKutf8}
\usepackage{caption}
\usepackage{capt-of}
\usepackage{amsmath}
\usepackage{CJKutf8}
\usepackage{amssymb}
\usepackage{pifont}
\usepackage{booktabs}
\usepackage{bbding}

\usepackage[T1]{fontenc}

\usepackage[utf8]{inputenc}

\usepackage{microtype}

\title{Towards Fine-grained Causal Reasoning and QA}

\author{
    Linyi Yang \textsuperscript{\rm1,2 }$^*$,
    Zhen Wang \textsuperscript{\rm3} $^*$,
    Yuxiang Wu \textsuperscript{\rm4},
    Jie Yang \textsuperscript{\rm3},
    Yue Zhang \textsuperscript{\rm1,2},
    \\
    \textsuperscript{1} School of Engineering, Westlake University \\
    \textsuperscript{2} Institute of Advanced Technology, Westlake Institute for Advanced Study \\
    \textsuperscript{3} Web Information Systems, Delft University of Technology \\
    \textsuperscript{4} University College London \\
    \texttt{\{yanglinyi, zhangyue\}@westlake.edu.cn} \\
    \texttt{\{Z.Wang-42\}@student.tudelft.nl},
    \texttt{\{j.yang-3\}@tudelft.nl}
    \\
}

\begin{document}
\maketitle

\def\thefootnote{*}\footnotetext{ These authors contributed equally to this work.}
\begin{abstract}

Understanding causality is key to the success of NLP applications, especially in high-stakes domains. Causality comes in various perspectives such as \emph{enable} and \emph{prevent} that, despite their importance, have been largely ignored in the literature. In this paper, we introduce a novel fine-grained causal reasoning dataset and present a series of novel predictive tasks in NLP, such as causality detection, event causality extraction, and Causal QA. Our dataset contains human annotations of 25K cause-effect event pairs and 24K question-answering pairs within multi-sentence samples, where each can contain multiple causal relationships. Through extensive experiments and analysis, we show that the complex relations in our dataset bring unique challenges to state-of-the-art methods across all three tasks and highlight potential research opportunities, especially in developing ``causal-thinking'' methods. 

\end{abstract}





\section{Introduction}

To entail a new goal of building more powerful AI systems beyond making predictions using statistical correlations \cite{ICLR21Counterfact,srivastava2020robustness,Li2021CausalBERTIC}, causality has received much research attention in recent years \cite{gao-etal-2019-modeling,scholkopf2021toward,feder2021causal,scherrer2021learning}. In NLP, understanding fine-grained causal relations between events in a document is essential in language understanding and is beneficial to various applications, such as information extraction \cite{gao-etal-2019-modeling}, question answering \cite{Chen2021FinQAAD}, and machine reading comprehension \cite{chen-etal-2021-probing}, especially in high-stakes domains such as medicine and finance.

\begin{figure}[t]
  \centering
  \includegraphics[width=.9\hsize]{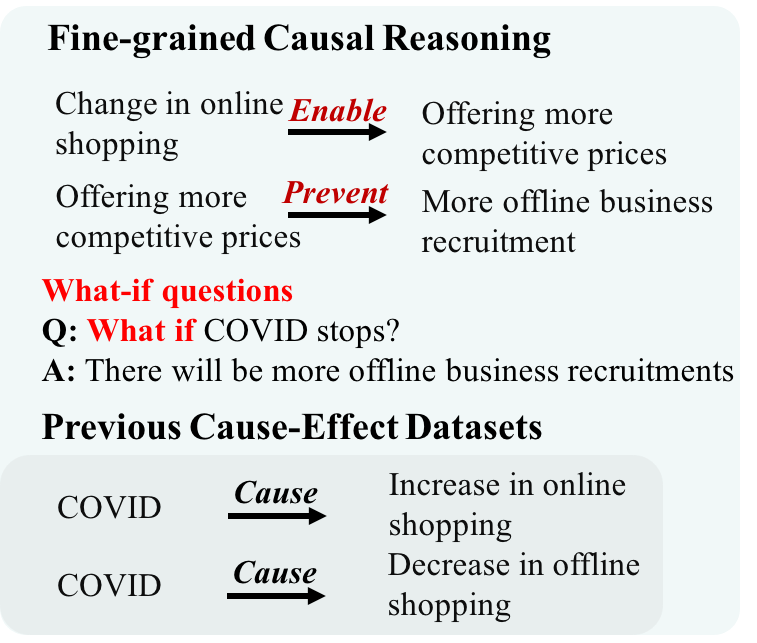}
  \caption{Comparisons between existing datasets and our dataset in the task of event causality analysis.}
  \label{fig:intro}
\end{figure}


Despite a large body of work on automatic causality detection and reasoning over text \cite{khoo1998automatic,mirza2014annotating,chang-chen-2019-word,fincausal}, relatively little work has considered the plethora of fine-grained causal concepts \citep{talmy1988force,wolff2005expressing}. For example, the spread of COVID-19 has \textit{led to} the boom in online shopping, (i.e., cause), but it also has \textit{deterred} ( i.e., prevent) people from going shopping-centres. Previous work has focused only on the ``cause'' relation. However, as suggested by the literature in classical psychology \citep{wolff2003models}, a single ``cause'' relationship cannot cover the richness of causal concepts in real-world scenarios; instead, it is important to understand possible fine-grained relationships between two events from different causal perspectives, such as \emph{enable} and \emph{prevent}. In practice, being able to recognize those fine-grained relationships can not only benefits the causal inference task \cite{keith2020text} but also facilities the construction of event evolutionary graph by providing more possible relation patterns between events \cite{li2018constructing}.

\begin{table*}[h]
\centering
\small
\begin{tabular}{l|ccccc}
\hline
\textbf{Datasets} & \begin{tabular}[c]{@{}l@{}}\textbf{Event}\\\textbf{Extraction} \end{tabular} & \begin{tabular}[c]{@{}l@{}}\textbf{Causal}\\\textbf{Reasoning} \end{tabular} & \begin{tabular}[c]{@{}l@{}}\textbf{Fine-grained}\\\textbf{Causality} \end{tabular} & \begin{tabular}[c]{@{}l@{}} \textbf{Span-based}\\\textbf{QA} \end{tabular} \\ \hline
FinCausal \cite{fincausal} & \CheckmarkBold & \CheckmarkBold & \XSolidBrush & \XSolidBrush \\
COPA \cite{copa} & \XSolidBrush & \CheckmarkBold  & \XSolidBrush  & \XSolidBrush \\ 
CausalBank \cite{ijcai2020causalbank} & \XSolidBrush & \CheckmarkBold  & \XSolidBrush  & \XSolidBrush \\ \hline
SQuAD \cite{squad2} &  \XSolidBrush &  \XSolidBrush & \XSolidBrush & \CheckmarkBold \\
LogiQA \cite{liu2020logiqa} &  \XSolidBrush &  \CheckmarkBold & \XSolidBrush & \XSolidBrush \\
DROP \cite{dua2019drop} & \XSolidBrush & \XSolidBrush &  \XSolidBrush & \CheckmarkBold \\
DREAM \cite{sun2019dream} & \XSolidBrush & \CheckmarkBold &  \XSolidBrush & \XSolidBrush \\
RACE \cite{lai2017race} & \XSolidBrush & \CheckmarkBold &  \XSolidBrush & \XSolidBrush \\\hline
FCR (Ours) &  \CheckmarkBold & \CheckmarkBold & \CheckmarkBold & \CheckmarkBold \\ \hline
\end{tabular}
\caption{Comparisons of our fin-grained causal reasoning dataset and related public datasets.}
\label{table:datasetcomparison}
\end{table*}

Motivated by \citet{wolff2003models}, we extend the causal reasoning task in NLP from a shallow ``cause'' relationship to three possible fine-grained relationships when constructing our dataset, including \emph{cause}, \emph{\textbf{enable}}, and \emph{\textbf{prevent}}. Formally, the \emph{\textbf{enable}} relationship can be expressed as the sufficient but not necessary condition between events, while the \emph{\textbf{cause}} relationship typically refers to the necessary and sufficient condition. Based on the hand-labeled, fine-grained, cause-effect pairs extracted from text corpus, we construct a fine-grained causal reasoning (FCR) dataset, which consists of 25, 193 cause-effect pairs as well as 24, 486 question-answering pairs, in which almost all questions are ``why'' and ``what-if'' questions concerning the three fine-grained causalities. 

To the best of our knowledge, FCR is the first human-labeled fine-grained event causality dataset. We define a series of novel tasks based on that, including \emph{causality detection}, \emph{fine-grained causality extraction}, and \emph{Causal QA}. In contrast to the performance of the state-of-the-art models on traditional cause-effect detection (e.g., FinCausal \cite{fincausal}) on the 94\% F1 score, experimental results show a significant gap between machine and human ceiling performance (74.1\% vs. 90.53\% accuracy) in our fine-grained task, providing the evidence that current statistical big models still struggle to solve the causal reasoning problem. 

\section{Related Work}

Table \ref{table:datasetcomparison} compares our dataset with datasets in the domain of both event causality and question answering (QA). In general, neither cause-effect detection datasets nor QA datasets considered fine-grained causal reasoning tasks. FinCausal \cite{mariko-etal-2020-financial} dataset is the most relevant to ours, which developed a relatively small dataset from the Edgar Database\footnote{\url{https://www.sec.gov/edgar/}} focusing on the simple ``cause'' relation only and do not contain QA tasks. We release our dataset and code at Github\footnote{\url{https://github.com/YangLinyi/Fine-grained-Causal-Reasoning}}.

\textbf{Fine-grained Causal Reasoning.} There is a deep literature on causal inference techniques using non-text datasets \cite{pearl2009causality,morgan2015counterfactuals,keith2020text,feder2021causal}, and a line of work focusing on discovering the causal relationship between events from textual data \cite{gordon2012semeval,mirza2016catena,du-etal-2021-excar}.Previous efforts lie on the graph-based event causality detection tasks \cite{tanon2017completeness,ijcai2020causalbank,du-etal-2021-excar} and the event-level causality detection tasks \cite{mariko-etal-2020-financial,fnp-2021-financial,Gusev2021HeadlineCauseAD}. However, causal reasoning for text data with a special focus on fine-grained causality between events has been relatively little considered. A contrast between our dataset and the previous causality detection dataset is shown in Fig. 1. As can be seen, given the same passage ``\textit{COVID-19 has accelerated change in online shopping, and given Amazon's ... it will result in economic returns for years to come and offering more competitive prices compared to an offline business that brings pressures for the offline business recruitment.}'', previous work can extract facts such as ``COVID-19 causes an increase in online shopping'', yet cannot detect the subsequence for Amazon to ``offer more competitive prices'', and further the negative influence on offline business recruitment, both of which can be valuable for predicting the future events.

\begin{figure*}[t]
  \centering
  \includegraphics[width=\hsize]{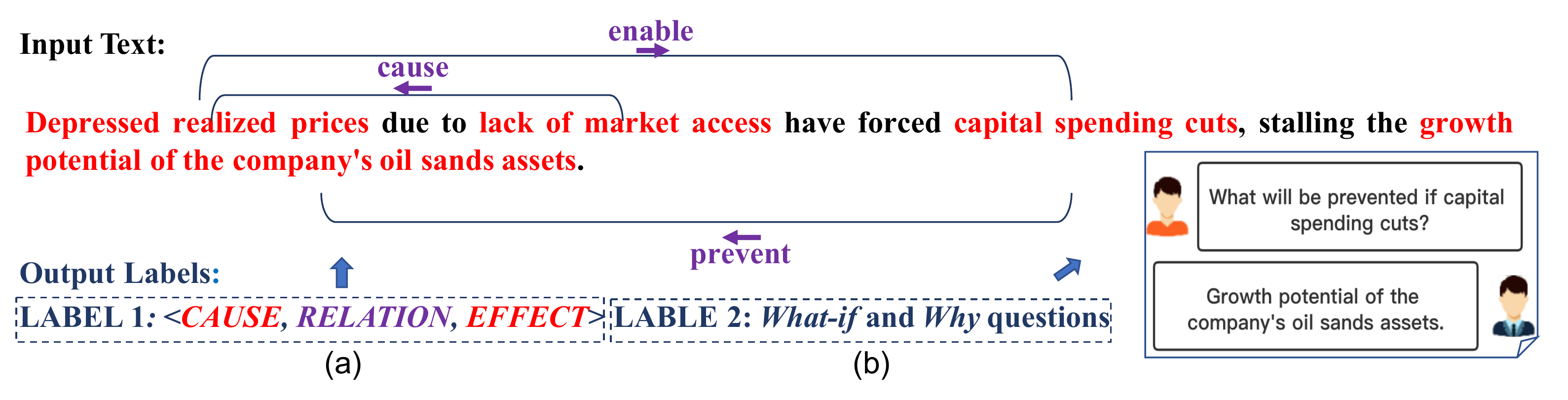}
  \caption{Illustration of our crowdsourcing tasks using an example that contains all three types of causal relationship.}
  \label{fig:sentence_example}
\end{figure*}

\textbf{Causal QA.}
Our QA task is similar to the machine reading comprehension setting \citep{huang2019cosmos} where algorithms make a multiple-choice selection given a passage and a question. Nevertheless, we focus on causal questions, which turn out to be more challenging. 

Existing popular question answering datasets \cite{sun2019dream,liu2020logiqa,mutual} mainly focus on \textit{what}, \textit{who}, \textit{where} and \textit{when} questions, making their usage scenarios somewhat limited. SQuAD \cite{squad1, squad2} consists of factual questions concerning Wikipedia articles, and some unanswerable questions are involved in SQuAD2.0. Although there are some datasets contain the causal reasoning tasks \cite{lai2017race,sun2019dream,mutual}, none of them consider answering questions by text span. Span-based question answering problems have gained wide interest in recent years \cite{yang2018hotpotqa,huang2019cosmos,lewis2021paq}. The answers in DROP \cite{dua2019drop} may come from different spans of a passage and require some combination technologies to get the correct answer.  Compared with these datasets, none of which has features of causal reasoning and span-based QA simultaneously, our dataset is the first to leverage fine-grained human-labeled causality for designing the Causal QA task consisting with ``Why'' and ``What-if'' questions.

\section{Dataset}

We collected an analyst report dataset from Yahoo Finance\footnote{We have received the written consent from the Yahoo Finance.}, which contains 6,786 well-processed articles between December 2020 and July 2021. Each instance corresponds to a specific analyst report on a U.S. listed company, which highlights the strengths and weaknesses of the company.

\subsection{Crowdsourcing}

The original FCR dataset consists of 6,786 articles in 54, 289 sentences. We employ editors from a crowd-sourcing company to complete several human annotation tasks. Several pre-processing steps required crowd-sourcing efforts were carried out to prepare the raw dataset, including (1) A binary classification task for the causality detection; (2) A span labeling task to mark the cause and effect formatted as text chunks (a given instance may contain multiple causal relations), and give event pairs a fine-grained causality label, including \emph{cause}, \emph{cause\_by}), \emph{enable}, \emph{enable\_by}), \emph{prevent}, \emph{prevent\_by}), and \emph{irrelevant} relation, where the suffix \emph{``\_by''} means the effect comes before its cause; (3) A re-writing task to generate the following question-answering dataset by using the labeled event triples.



\textbf{Causality Detection.} We first focus on a binary classification task of the causality detection, as such, removed sentences with outcome types of non-causal relationships, leaving only those text sequences (one or two sentences) that are considered containing at least a causal relation. 

\textbf{Fine-grained Event Causality.} Given the sentences each containing at least one event causality, human annotators are required to highlight all the event causalities and give each instance a fine-grained label. As shown in Fig. \ref{fig:sentence_example}(a), a single sentence can have more than one event causality, which will be stored as triples containing $\langle$ \textit{cause, relation, effect} $\rangle$.

\textbf{Causal QA.}  As shown in Fig. 2(b), we design a novel and challenging causal reasoning QA task based on the fine-grained causality labels. We expand each \(<cause, relation, effect>\) triple for generating a plausible question-answer pair. Different templates have been designed for different types of questions. For example, the active causal relations -- \emph{CAUSE}, \emph{ENABLE}, and \emph{PREVENT} -- could usually be used for generating why-questions while the corresponding passive causal relations could be used for generating what-if questions. 

\begin{table}[t]
\centering
\small
\begin{tabular}{ll}
\hline
\textbf{Metric} & \textbf{Counts} \\ \hline
\multicolumn{2}{c}{Causality Sentence Classification (\textbf{Task1})} \\ \hline
\#Positive Instances & 21,046 \\
\#Negative Instances & 29,979 \\
\#Multi-sentence Samples & 846 \\
\#Average Token Length of POS Samples & 42.8 \\ 
\#Average Token Length of NEG Samples & 41.3 \\ \hline
\multicolumn{2}{c}{Cause-Effect Event Pairs (\textbf{Task2})} \\ \hline
\#Causal Text Chunks & 45, 710 \\
\#Uni-causal Text Spans & 18, 457 \\
\#Multi-causal Text Spans & 3, 017 \\
\#Average Token Length of Cause Spans & 16.0 \\ 
\#Average Token Length of Effect Spans & 15.2 \\ \hline
\multicolumn{2}{c}{Causal QA Pairs (\textbf{Task3})} \\ \hline
\#Total Number of QA pairs & 24, 486 \\
\#Average Token Length of Context & 191.7 \\
\#Average Token Length of Questions & 20.1 \\ 
\#Average Token Length of Answers & 15.4 \\
\#Variance of Answer Length & 69.15 \\\hline
\end{tabular}
\caption{Statistics for causality detection, cause-effect pairs and QA pairs.}
\label{table:statistics}
\end{table}
\textbf{Quality Control.} To ensure high quality, we restricted the participants to experienced human labelers with relevant records. For each task, we conducted pilot tests before the crowd-sourcing work officially began, receiving feedback from quality inspectors and revising instructions accordingly. We filter out the sentences regarding the estimation of the stock price movement due to the naturally high-sensitive features and uncertainty of the complex financial market. After the first-round annotation (half of the data), we manually organized spot checks for 10\% samples in the dataset and revised the incorrect labels. After review, we revised roughly 3\% of instances and refused the labelers with above 10\% error rate from participating in the second-round data annotation. Finally, the inter-annotators agreement (\textbf{IAA}) ratio is 91\% for fine-grained causality labels, and the F1 score of the inter-annotators agreement (\textbf{IAA}) ratio is 94\% for causal question-answer pairs.

Finally, we obtained a dataset of 51,025 instances (21,046 contain at least one causal relation) with fine-grained labels of cause-effect relations that were subsequently divided into training, validation, and testing sets for the following experiments. It may be worth noting that we sort the dataset in chronological order because the future data is not expected to be used for predictions.

\subsection{Discussion}
The primary data statistic of the FCR dataset is shown in Table \ref{table:statistics} for three different tasks. We observe that there is no significant difference in the average token numbers between positive and negative examples for Task 1 and 2, which shows that predictive models are difficult to learn from shortcut features \cite{sugawara2018makes,sugawara2020assessing,lai2021machine} (e.g., the instance length) during the training process. Furthermore, our dataset contains 846 multi-sentence samples, and 3,017 text chunks contain more than one causal relation in one instance, which requires a complex reasoning process to get the correct answer, even for a human. Most importantly, unlike other QA datasets \citep{sugawara2018makes,sugawara2020assessing} that can easily benefit from the test-train overlap as revealed by \citep{lewis2021question,liu2021challenges,wang-etal-2021-generative}, our dataset is sorted in chronological order so that the future test data could be theoretically difficult to coincide with the training set. This allows us to obtain greater insight into what extent models can actually generalize.

\begin{figure}[t]
\centering
\subfigure[Companies Count]{
\begin{minipage}[t]{0.5\linewidth}
\centering
\includegraphics[width=1.5in]{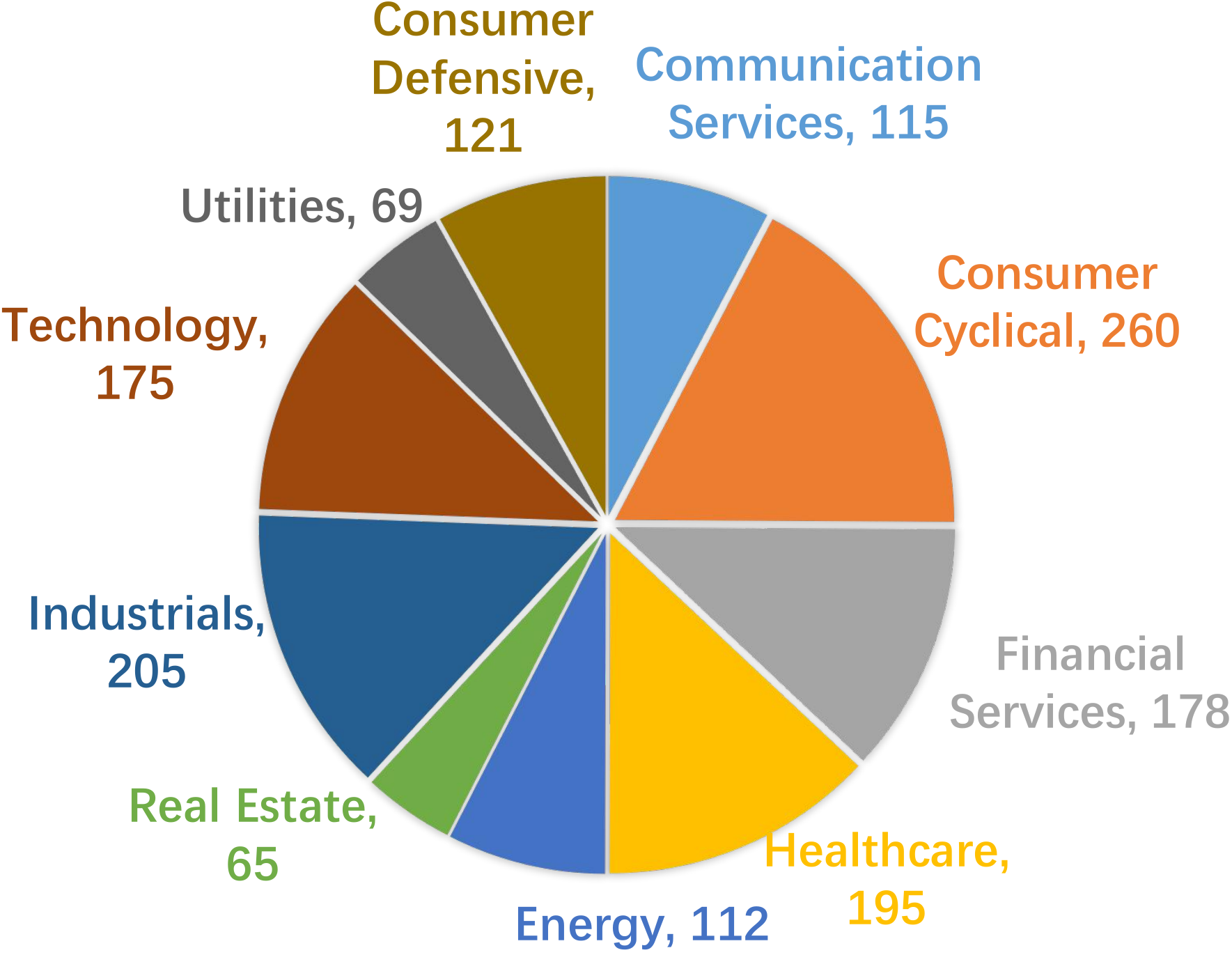}
\end{minipage}%
}%
\subfigure[Reports Count]{
\begin{minipage}[t]{0.5\linewidth}
\centering
\includegraphics[width=1.5in]{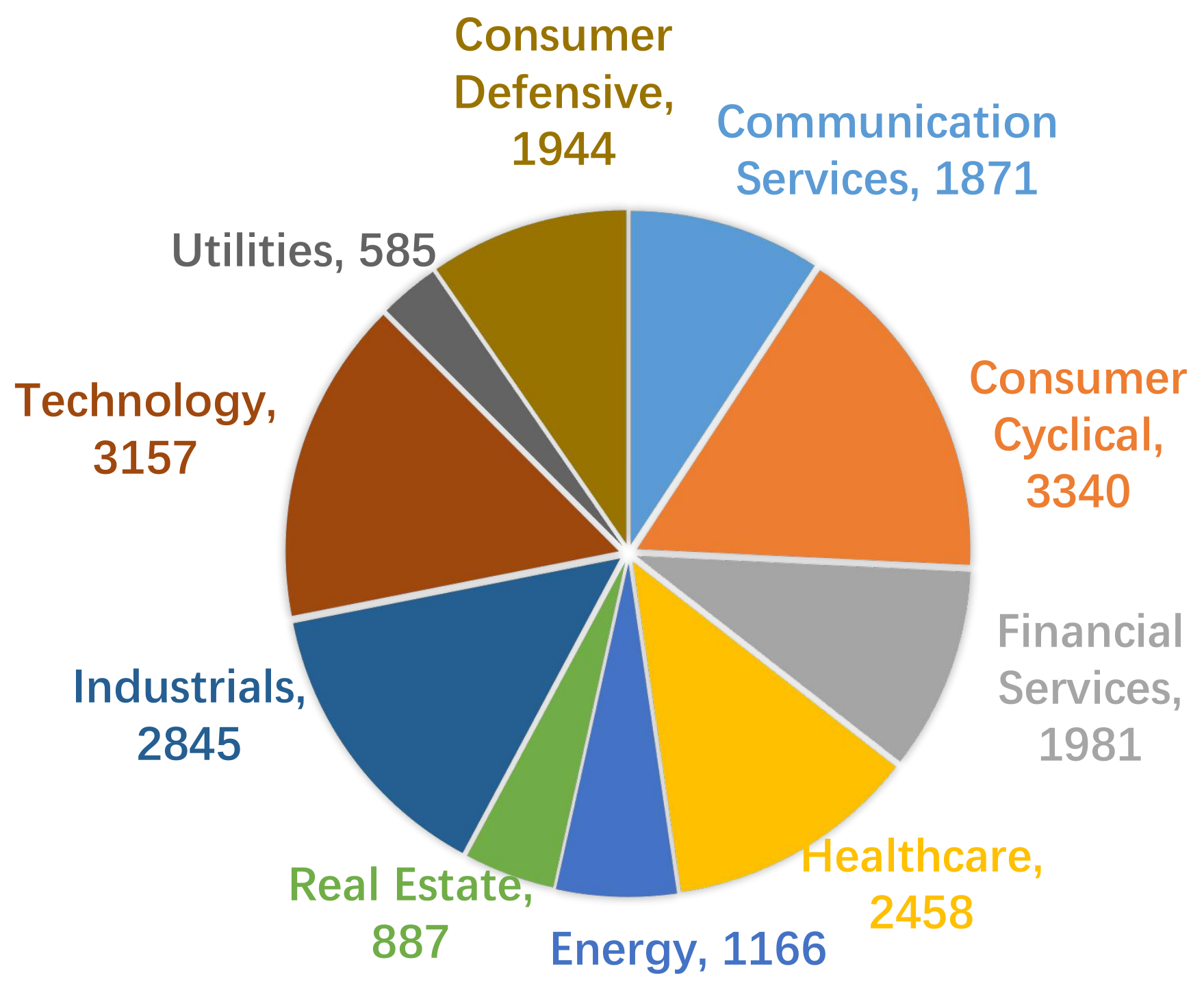}
\end{minipage}}
\centering
\caption{Sector distributions on companies and reports.}
\label{fig:sector}
\end{figure}

\begin{figure*}[t]
  \centering
  \includegraphics[width=.9\hsize]{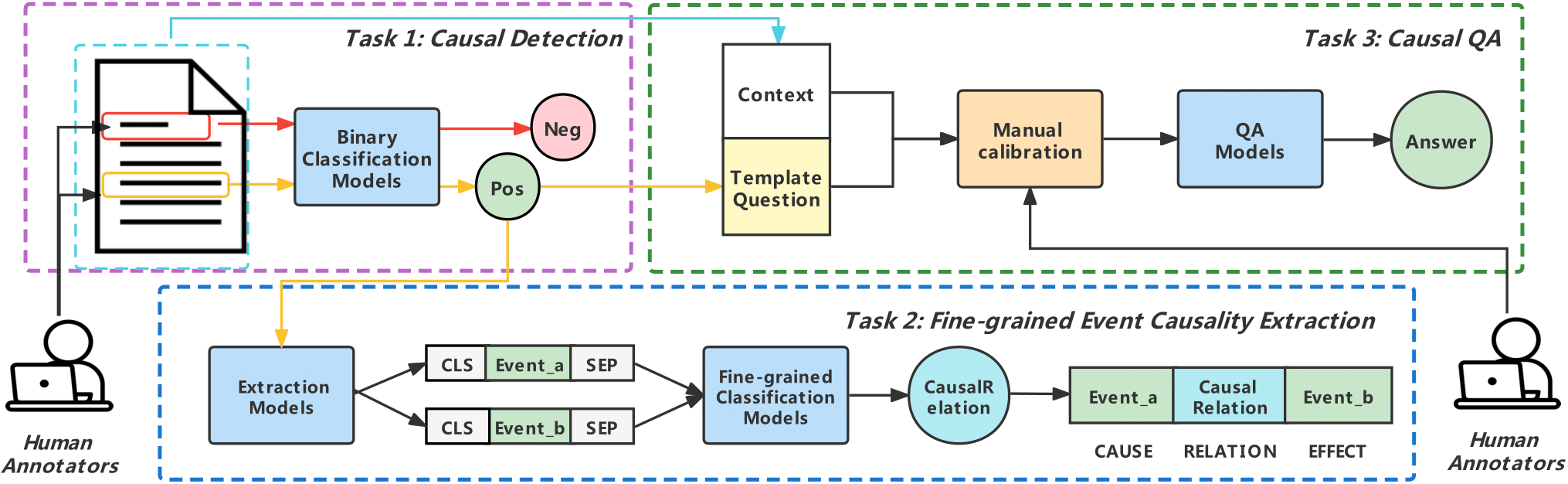}
  \caption{The pipeline of experiments designed for fine-grained causal reasoning in NLP.}
  \label{fig:pipeline}
\end{figure*}

\subsection{Meta-information}

Our dataset contains multi-sentence instances with fine-grained causality labels and the meta-information (company names and published dates). We list the sector distribution information in \ref{fig:sector}, in order to show that our dataset contains cause-effect pairs from different domains, although it has been collected from the single source — financial reports. The top three largest sectors belong to Consumer Cyclical, Industrial, and Technology. While instances from Utilities companies are the smallest group in our dataset. The use of the meta-information is two-fold. First, we choose the top three largest domains to perform out-of-domain evaluations (see Appendix B). Second, company names would be used for generating question templates. 


\section{Tasks and Methods}

The pipeline of our experiments is shown in Fig. \ref{fig:pipeline}.  We define three tasks on our FCR dataset and build strong benchmark results for each task. 

\subsection{Causal Detection}

First, as a prerequisite, models are evaluated on a binary classification task to predict whether a given text sequence contains a causal relation (\textbf{Task 1}). Second, we set up a joint event extraction and fine-grained causality task for identifying text chunks describing the cause and effect, respectively, and which fine-grained causality category it belongs to (\textbf{Task 2}). Finally, we design a question answering task for answering the challenging ``why-questions'' and ``what-if questions'' (\textbf{Task 3}). 

The dataset consists of instances labeled with positive for the binary classification task if a given instance contains one causal relation and negative non-causal instances. The input data is extracted from the raw dataset directly, which contains 846 multi-sentence samples. We include multi-sentence samples besides a single sentence because causality could be found in multi-sentence contexts. 


\subsection{Fine-grained Event Causality Extraction}

In Task 2, we use [CLS] and [SEP] to mark the event's begin and end positions, respectively. For example, we have ``[CLS] Better card analytics, increased capital markets and M\&A offerings, and bolt-on acquisitions [SEP] should help drive [CLS] growth in fee income [SEP]'', in which ``Better card analytics, increased capital markets and M\&A offerings, and bolt-on acquisitions'' is a ``cause'' event and ``growth in fee come'' is an ``effect'' event. In total, there are 33, 634 event triples with seven-class labels used for our experiments.

Then, we conduct the cause-effect extraction between samples. We consider cause-effect extraction a multi-span event extraction task as complex causal scenarios containing multiple causes or effects within a single instance is under consideration. We set the label for the first token of a cause or effect to ``B'', the rest of the tokens within the detected text chunks are given the label ``I'', and the other words in a given instance are set to ``O''. The results of event causality extraction are reserved for generating causal questions.

\subsection{Causal QA}

Both extractive methods and generative methods have been evaluated. For the extractive QA task, we adopt the same methods as the previous Transfomer-based QA works \cite{binary2020qa}. In particular, we first convert the context \(C = \left(c_{1}, c_{2}, ..., c_{l}\right)\) and question \(Q=\left(q_{1}, q_{2}, ..., q_{l^{\prime}}\right)\) into a single sequence \(X= [CLS]\) \(c_{1} c_{2} \ldots c_{l}\)  \([SEP]\) \(q_{1} q_{2} \ldots q_{l^{\prime}}[\mathrm{SEP}]\), passing it to the pre-trained Transformer encoders for predicting the answer span boundary (start and end). 

\subsection{Models}
We consider using both classical deep learning models -- CNN-Test \cite{kim2014convolutional} and HAN \cite{yang2016hierarchical}  -- and Transformer-based models downloaded from Huggingface\footnote{\url{https://github.com/huggingface/models}} -- BERT \cite{bert}, RoBERTa \cite{roberta}, and SpanBERT \cite{Joshi2020SpanBERTIP} -- as predictive models. 

In addition, we perform a causal reasoning QA task by leveraging six Transformer-based pre-trained models built on the recently advanced Transformer architectures \cite{transformer} with the framework provided by Huggingface\footnote{\url{https://github.com/huggingface/transformers}}, including BERT-base, BERT-large, RoBERTa-base, RoBERTa-large, RoBERTa-base-with-squad, and RoBERTa-large-with-squad\footnote{\url{https://huggingface.co/navteca/roberta-large-squad2}}. Furthermore, pre-trained seq2seq models such as T5 \cite{t5}, or BART \cite{bart} are fine-tuned on QA-pairs as the benchmark methods of the generative QA tasks. In particular, we consider T5-small, T5-base, T5-large, BART-base, and BART-large models for building the benchmark results.

\begin{table}[t]
\centering
\small
\begin{tabular}{cllll}
\hline
\multirow{2}{*}{\textbf{Methods}} & \multicolumn{2}{c}{\textbf{Dev.}} & \multicolumn{2}{c}{\textbf{Test}} \\
 & F1 & Acc & F1 & Acc \\
\hline
CNN-Text & 81.35 & 81.59 & 80.03 & 81.01 \\
HAN & 81.18 & 81.23 & 80.60 & 81.26 \\ \hline
BERT-base & 83.72 & 84.23 & 84.02 & 84.43 \\
BERT-large & 84.03 & 84.41 & 84.63 & 84.90 \\
SpanBERT-base & 84.09 & 84.38 & 84.51 & 84.72 \\
SpanBERT-large & 84.43 & 84.80 & 84.55 & 84.82 \\ 
RoBERTa-base & \textbf{84.59} & \textbf{85.16} & 84.31 & 84.76 \\
RoBERTa-large & 84.39 & 84.75 & \textbf{84.64} & \textbf{84.89} \\\hline
Human & \multicolumn{1}{c}{-} & \multicolumn{1}{c}{-} & 94.32 & 95.94 \\ \hline
\begin{tabular}[c]{@{}c@{}} Best Results \\ $@$FinCausal \end{tabular}  & \multicolumn{1}{c}{-} & \multicolumn{1}{c}{-} & 97.75 & 97.76 \\ \hline
\end{tabular}
\caption{The results of the causal sentence classification. 'F1' refers to the Macro F1.}
\label{table:binarycls}
\end{table}

\begin{table*}[t]
\centering
\small
\begin{tabular}{cllllllllllll}
\hline
\multirow{2}{*}{\textbf{Model}} & \multicolumn{4}{c}{\textbf{Event Causality Extraction}} & \multicolumn{4}{c}{\textbf{Fine-grained Classification}} &  \multicolumn{2}{c}{\textbf{Joint Evaluation}} \\
 & \multicolumn{2}{c}{\textbf{Dev.}} & \multicolumn{2}{c}{\textbf{Test}} & \multicolumn{2}{c}{\textbf{Dev.}} & \multicolumn{2}{c}{\textbf{Test}} & \multicolumn{1}{c}{\textbf{Dev.}} & \multicolumn{1}{c}{\textbf{Test}} \\
 & F1 & EM & F1 & EM & F1 & ACC & F1 & ACC & EM & EM \\
\hline
BERT-base & 84.37 & 51.48 & 85.30 & 53.53 & 71.74 & 70.43 & 63.72 & 71.72 & 21.21 & 20.15 \\
BERT-large & 85.13 & 50.34 & 86.93 & 52.88 & 70.90 & 64.16 & 60.24 & 69.85 & 17.54 & 21.73 \\
RoBERTa-base & 85.67 & 53.41 & 86.32 & 56.04 & 73.09 & 68.37 & 65.99 & 71.63  & 20.45 & 19.08 \\
RoBERTa-large & 85.12 & 54.70 & 85.95 & 56.77 & \textbf{74.54} & \textbf{71.99} & \textbf{68.99} & \textbf{74.09} & 20.46 & 19.77\\
SpanBERT-base & \textbf{85.84} & 55.40 & \textbf{86.82} & 57.26 & 71.18 & 68.40 & 63.73 & 70.52  & 21.17 & 21.09 \\
SpanBERT-large& 85.50 & \textbf{57.40} & 86.33 & \textbf{60.26} & 73.65 & 68.15 & 64.43 & 72.93  & \textbf{23.01} & \textbf{21.78} \\\hline 
Human Performance & \multicolumn{1}{c}{-} & \multicolumn{1}{c}{-} & 94.32 & 81.34 & \multicolumn{1}{c}{-} & \multicolumn{1}{c}{-} & 88.61 & 90.53 & \multicolumn{1}{c}{-} & \multicolumn{1}{c}{-} \\ \hline
\end{tabular}
\caption{The results of the joint event causality detection (task2), 'F1' refers to the Macro F1. 'ACC.' is short for the accuracy, 'EM' refers to exact match and spe.}
\label{table:task2result}
\end{table*}

\section{Experiments}

We present and discuss the results of \textbf{Task 1-3} based on FCR dataset in this section. 

\subsection{Settings}

For hold-out evaluation, we split our dataset into mutually exclusive training/validation/testing sets in the same ratio of 8:1:1 for all tasks. Predictive models and data splitting strategies have been kept the same among these tasks for building the benchmark results of each task. In line with the best practice, model hyper-parameters are tuned using the validation set. Both validation results and testing results will be reported in experiments. We use Adam as the optimizer and adopt the trick of decay learning-rate with the steps increase to train our model until converging for all models. The Macro F1-score and accuracy are used for evaluating the event causality analysis task, and the exact match and F1-score are used for Causal QA. 

The Macro F1-score is defined as the mean of label-wise F1-scores:
\begin{equation}
    \text { Macro F1-score }=\frac{1}{N} \sum_{i=0}^{N} \mathrm{~F} 1 \text {-score }_{i}
\end{equation}
where \(i\) is the label index and \(N\) is the number of classes.

\subsection{Causality Detection}

The causal detection result is shown in Table \ref{table:binarycls}. We find that although Transformer-based methods achieve much better results than other methods -- CNN and HAN using ELMO embeddings -- on judging whether an instance contains at least a causal relationship (RoBERTa-Large can get the highest F1 Score -- 84.64), it is still significantly below the human performance (84.64 vs. 94.32). The results of human performance are reported by quality inspectors from the crowdsourcing company. It is worth noting that the best results on the FinCausal \cite{fincausal} dataset can reach the human-level result (F1 = 97.75), providing indirect evidence that our dataset is more challenging caused by more complex causality instances.

\subsection{Fine-grained Event Causality Extraction}

\begin{table}[t]
\centering
\small
\begin{tabular}{llllll}
\hline
\multirow{2}{*}{\textbf{Category}} & \multirow{2}{*}{\textbf{Counts}} & \multicolumn{2}{c}{\textbf{Dev.}} & \multicolumn{2}{c}{\textbf{Test}} \\
&  & F1 & Acc & F1 & Acc \\ \hline
Irrelevant & 8,441 & 84.17 & 86.49 & \textbf{84.40} & \textbf{85.55} \\
Cause & 8,428 & 73.60 & 73.93 & 74.00 & 76.61  \\
Cause\_By & 7,437 & 80.21 & 84.94 & 79.62 & 83.69 \\
Enable & 5,506 & 63.42 & 60.91 & 62.61 & 58.70 \\ 
Enable\_By & 2,367 & 47.66 & 41.06 & \textit{41.49} & \textit{35.68} \\ 
Prevent & 1,086 & 79.79 & 71.43 & 76.34 & 67.87 \\ 
Prevent\_By & 369 & 55.88 & 52.78 & 64.46 & 65.00 \\ \hline
\end{tabular}
\caption{Error analysis for fine-grained classifications.}
\label{table:finegrainederror}
\end{table}

The results of the fine-grained event causality extraction task are shown in Table \ref{table:task2result}. We find that SpanBERT and RoBERTa model can achieve the best performance for event causality extraction (F1 = 86.82 and EM = 60.26) and fine-grained classification (F1 = 68.99 and EM = 74.09), respectively. Nevertheless, all methods perform dramatically worse on the more challenging joint task, where the prediction is judged true only if event extraction and classification results exactly match the ground truth. Although the SpanBERT-large model can achieve the highest 21.78 EM on the test set, there is still much room for improvement. 

We find that the large Transformer-based models \cite{transformer} with larger parameter sizes could not improve the performance on these tasks based on the FCR dataset by comparing the test performance of BERT-base (\textbf{63.72} in F1, \textbf{71.72} in ACC) with BERT-large (60.24 in F1, 69.85 in ACC) on the task of the fine-grained classification. It sheds new light that increasing the parameter size could not be helpful for causal reasoning tasks. 

A more detailed error analysis by using the best-performing RoBERTa-Large model is given in Table \ref{table:finegrainederror}. The model performs relatively better in terms of the F1 score when predicting simple causal relations -- Irrelevant (84.40), Cause (74.00), Cause\_by (79.62), and Prevent (76.34), but worse on predicting complex relations -- Enable (62.61) and Enable\_by (41.49) -- and the category with few examples -- Prevent\_by (64.46). This indicates that fine-grained causal reasoning brings the unique challenge for pre-trained models.

\newcommand\boldblue[1]{\textcolor{blue}{\textbf{#1}}}
\newcommand\boldred[1]{\textcolor{red}{\textbf{#1}}}
\newcommand\boldgreen[1]{\textcolor{green}{\textbf{#1}}}
\newcommand\boldorange[1]{\textcolor{orange}{\textbf{#1}}}

\begin{table}[t]
\centering
\small
\begin{tabular}{lllll}
\hline
 & \multicolumn{2}{c}{\textbf{Dev.}} & \multicolumn{2}{c}{\textbf{Test}} \\
\multirow{-2}{*}{\textbf{Causal QA}} & F1 & EM & F1 & EM \\
\hline
BERT-base & 79.90 & 55.52 & 79.33 & 55.70 \\
BERT-large & 82.48 & 59.24 & 82.37 & 58.71 \\
RoBERTa-base & 82.96 & 60.13 & 83.11 & 60.33 \\
\textit{SQuAD2.0-only} & \textit{64.87} & \textit{26.71} & \textit{65.20} & \textit{27.36} \\
SQuAD2.0-enhanced & 84.39 & 61.22 & 84.34 & 61.17 \\
RoBERTa-large & 84.28 & \textbf{61.69} & 84.35 & \textbf{61.76} \\
\textit{SQuAD2.0-only} & \textit{63.99} & \textit{26.02} & \textit{63.82} & \textit{25.26} \\
\textbf{SQuAD2.0-enhanced} & \textbf{84.65} & 61.63 & \textbf{84.65} & 61.58 \\
\hline
\multicolumn{5}{c}{Generative Methods} \\
\hline

BART-base & 74.34 & 35.81 & 74.35 & 36.16 \\
T5-small & 75.98 & 42.31 & 76.40 & 41.61 \\
\textbf{T5-Large} & \textbf{81.95} & \textbf{48.17} & \textbf{81.77} & \textbf{47.43} \\\hline
\end{tabular}
\caption{The results of causal reasoning QA using both extractive methods and generative methods. 'SQuAD2.0' refers to the evaluation results using the model trained with the training set of SQuAD2.0\footnote{https://rajpurkar.github.io/SQuAD-explorer/} only.}
\label{tab:qatest}
\end{table}

\begin{table*}[ht]
\centering
\small
\begin{tabular}{llccc}
\hline
\textbf{Dataset}  & \textbf{Method} & \textbf{F1} & \textbf{ACC} & \textbf{EM} \\ \hline
SQuAD1.1 \cite{squad1} & LUKE \cite{yamada2020luke} & 95.7  & - & 90.6 \\
SQuAD2.0 \cite{squad2} & IE-Net \cite{gao2019intra} & 93.2  & - & 90.9 \\
DROP \cite{dua2019drop} & QDGAT \cite{chen2020question} & 88.4  & - & - \\
HotpotQA \cite{yang2018hotpotqa} & BigBird-etc \cite{zaheer2020big} & 95.7  & - & 90.6 \\\hline
\textbf{Reasoning Based Datasets}  &  & & & \\ \hline
LogiQA\cite{liu2020logiqa} & DAGAN \cite{huang2021dagn} & -  & 39.3 & - \\
Causal QA (Ours) & RoBERTa-SQuAD & 84.7 & 85.6 & 61.6 \\\hline
\end{tabular}
\caption{The comparison of best performance between our dataset and other popular QA datasets.}
\label{qasotacomparison}
\end{table*}

\subsection{Causal QA}
We provide both quantitative analysis and qualitative analysis for Causal QA. In addition, we compare the best performance on our dataset and other popular QA datasets.

\subsubsection{Quantitative Analysis}


The results of Causal QA are given in Table \ref{tab:qatest}, where the bold values indicate the best performance while the italic values show the results of transfer learning methods trained by the SQuAD2.0 training data only. We find that the best-performing generative model -- T5-Large -- can achieve comparable results with the RoBERTa-large in terms of the F1 (81.77 vs. \textbf{84.35}). Meanwhile, the average EM of generative methods is largely below the extractive methods using the same training data. Second, the results of models trained with SQuAD2.0 data are much worse than those models trained with the original FCR training set in terms of the F1 score (65.20 vs. \textbf{83.11} for RoBERTa-base and 63.82 vs. \textbf{84.35} for RoBERTa-large). On the other hand, we note a distinct improvement of using SQuAD2.0 data for initially training for both RoBERT-base (from 83.11 to \textbf{84.34}) and RoBERTa-large (from 84.35 to \textbf{84.65}), which indicates that the training with additional well-labelled data could bring significant benefits for Causal QA. This may hint that the current QA data sources are still helpful for improving the performance of the causal reasoning QA task, although further research is required, as to what extent models can actually benefit from the additional data for the generalization is hard to be evaluated.

\begin{table*}[t]
\centering
\small
\begin{tabular}{p{0.41\textwidth}p{0.19\textwidth}p{0.16\textwidth}p{0.13\textwidth}}
\hline
\textbf{Context} & \textbf{Question} & \textbf{Gold Answer} & \textbf{Output Answer} \\ \hline

(Relation: Cause) Amazon's 2017 purchase of Whole Foods remains a threat ... The COVID-19 outbreak has lifted near-term revenue as shoppers spend more time at home. & Why \textcolor[rgb]{0.09, 0.45, 0.27}{\textbf{the COVID-19 outbreak has lifted near-term revenue}} for \boldorange{Amazon}? & \boldred{Shoppers spend more time at home} & Shoppers spend more time at home \\ \hline

(Relation: Enable) As a first mover in the local-market daily deals space, Groupon has captured a leadership position, but not robust profitability. & What enable \boldorange{Groupon} \textcolor[rgb]{0.09, 0.45, 0.27}{\textbf{capture a leadership position}}? & \boldred{A first mover in the local-market daily deals space} & A first mover in the local-market daily deals space\\
\hline

(Relation: Prevent\_By) In neurology, RNA therapies can reach their intended targets via intrathecal administration into spinal fluid, directly preventing the production of toxic proteins & What will be prevented if \boldred{intrathecal administration into spinal fluid}? & \textcolor[rgb]{0.09, 0.45, 0.27}{\textbf{The production of toxic proteins}} & The production of toxic proteins \\\hline

\textbf{Examples of Incorrect Predictions} & & & \\\hline
(Relation: Enable\_By) ... Through analyzing the data and applying artificial intelligence, the advertisers can improve the efficiency of advertisements through targeted marketing for Tencent ...  & What can help \boldred{\textbf{advertisers to improve the efficiency of advertisements}}? &  \textcolor[rgb]{0.09, 0.45, 0.27}{\textbf{Analyzing the data and applying artificial intelligence}} & Targeted marketing \\ \hline
(Relation: Cause\_By) Given expectations for more volatile equity, as well as some disruption as Brexit moves forward, it remains doubtful that flows will improve, a negative 3\%-5\% annual organic growth...  & Why \boldred{\textbf{a negative 3\%-5\% annual organic growth}} happened? & \textcolor[rgb]{0.09, 0.45, 0.27}{\textbf{Given ... as well as some disruption as Brexit moves forward}} & It remains doubtful that flows will improve. \\ \hline
\end{tabular}
\caption{Qualitative analysis of ``Why'' and ``What-if'' questions answering tasks based on the best-performed RoBERTa-Large model. The \boldorange{company name} can be found in the meta-information of our dataset. \boldred{Cause} and \textcolor[rgb]{0.09, 0.45, 0.27}{\textbf{Effect}} are extracted from the original context. The inputs of models consist with the context and question.}
\label{tab:qaerroranalysisi}
\end{table*}


\subsubsection{Challenges by Causal QA}


We are interested in better understanding the difficulty of the Causal QA task compared to other popular datasets regarding prediction performance. We list the best-performing model of several popular datasets in Table \ref{qasotacomparison}. In general, we find that reasoning-based tasks are more complex than other tasks in terms of the relatively low accuracy achieved by the state-of-the-art method. LogiQA is more challenging than our dataset (39.3 vs. 85.6 in accuracy) because it requires heavy logical reasoning rather than identifying causal relations from text. Moreover, we find that the state-of-the-art result on our dataset (RoBERTa-SQuAD) is dramatically worse than the best performance on other datasets (EM = 90.9 on SQuAD2.0 while EM = 61.6 on Causal QA). This may suggest that the model tends to output the partially right answer but fails to output the utterly correct answer, although further research is required, as the model still could be easily perturbed by the length of an event. Meanwhile, the human performance is still ahead of the best-performing model's result in the causal reasoning QA task. Thus, we argue that Causal QA is worth investigating by using more ``\emph{causal-thinking}'' methods in the future.

\subsubsection{Error Analysis}

Table \ref{tab:qaerroranalysisi} presents a qualitative analysis for Causal QA, where we highlight the question and answer parts extracted from the raw context. Human labelers label the gold answers while the BERT-based model generates the output answers.  The first two questions are answered correctly, while the last two instances show two typical patterns prone to errors. In the first incorrect example, the model outputs ``\textit{targeted marketing}'' using the keyword ``\textit{through}'' but fails to give the gold answer ``\textit{analyzing the data and applying artificial intelligence}''. This could be because the model fails to identify the difference between the same word appearing in two different positions. The last example shows that the model tends to output the answer closer to the question in the context instead of observing the whole sentence. The real reasons -- ``\textit{equity and credit markets}'' and ``\textit{Brexit}'' -- are ignored as it is relatively away for the question position.



\section{Conclusion}

We explored the efficacy of current state-of-the-art methods for causal reasoning tasks by considering a novel fine-grained reasoning setting and developing a dataset with rich human labels. Experimental results using the state-of-the-art pre-trained language models provide the evidence that there is much room for improvement on causal reasoning tasks, and a need for designing better solutions to correlation discovery related to event causality analysis and Why/What-if QA tasks.

\section{Ethical Statement}

This paper honors the ACL Code of Ethics. Public available financial analysis reports are used to extract fine-grained event relationships. No private data or non-public information was used. All annotators have received labor fees corresponding to their amount of annotated corpus. The code and data are open-sourced under the Creative Commons Attribution-NonCommercial-ShareAlike (CC-BY-NC-SA) license.

\bibliography{anthology,custom}
\bibliographystyle{acl_natbib}

\clearpage

\appendix

\section{Appendix: Annotation Instructions}

\begin{figure}[h]
  \centering
  \includegraphics[width=\hsize]{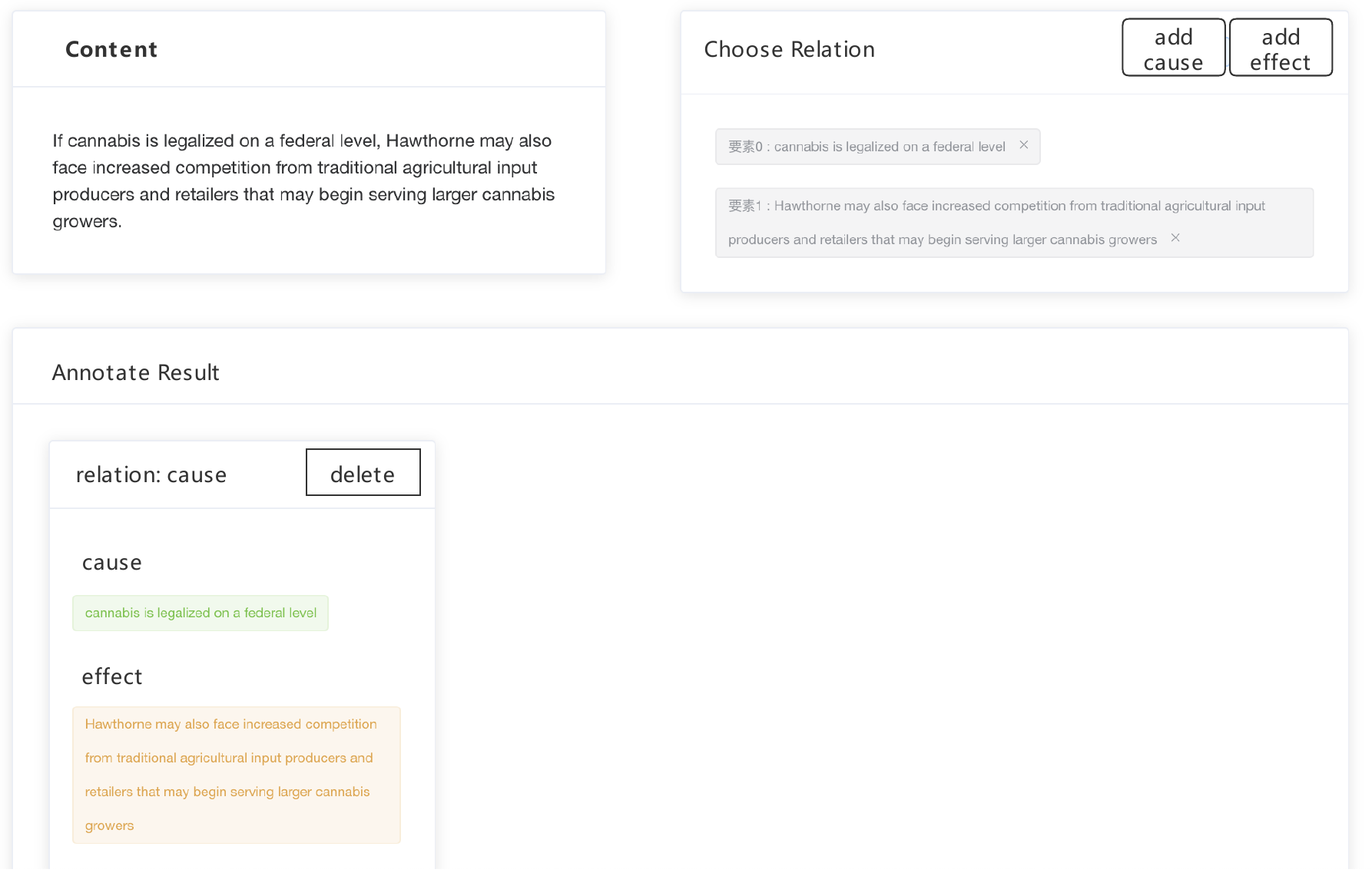}
  \caption{The annotation platform provided the crowdsourcing company for collecting annotations for fine-grained causality reasoning and Causal QA.}
  \label{fig:annotation}
\end{figure}

The annotation platform used in this work is introduced in Fig. \ref{fig:annotation}. As follows, we provide the detailed annotation instructions used for training the human labelers. Also, we show the annotation of some real examples stored in our dataset.

\subsection{General Instruction}

This is a annotation task related to event causality. In this task, you are asked to find all the cause-effect pairs and the fine-grained event relationship types from give passages.

\subsection{Steps}
\begin{enumerate}
    \item Please read your assigned examples carefully.
    \item A sentence is considered contains event causality if at least two events occur in it and the two events are causally related.
    \item If a sentence contains causality, mark it as \boldred{Positive}; otherwise, mark it as \textcolor[rgb]{0.09, 0.45, 0.27}{\textbf{Negative}}.
    \item For the \boldred{positive} sentence, first find all the events that occur in the sentence, and then pair the events to see if they constitute a causal relationship. The relationship much be one of \textbf{Cause}, \textbf{Enable} and \textbf{Prevent}.
    \item ``A causes B'' means B always happens if A happens. ``A enables B'' means A is a possible way for B to happen, but not necessarily. ``A prevents B'' means A and B cannot happen at the same time.
    \item Remember to annotate all event causality pairs. If there is no more pairs, process to the next passage.
\end{enumerate}

\subsection{Examples}

Here are some annotation examples, please read it before starting your annotation.

\textbf{Example 1:} Moreover, we do not think that DBK's investment banking operation has the necessary scale and set-up to outcompete peers globally or within Europe.

\textbf{Answer:} \# \textcolor[rgb]{0.09, 0.45, 0.27}{\textbf{Negative}}

\textbf{Explanation: } This is a sentence that contains no causal relationship between events.

\textbf{Example 2:} In our view, \boldorange{customers are likely to stay with VMware} because of \boldorange{knowledge of its product ecosystem as well as the risks and complexities associated with changing virtual machine providers}.

\textbf{Answer:} \# \boldred{Positive}

\textbf{Explanation:} This is a causal sentence. There exist two events marked by \boldorange{yellow} color. You should first annotate the two events and then give them the label according to their relationship, using one of \textbf{Cause}, \textbf{Enable} and \textbf{Prevent}. Here the relationship is \textbf{Cause}.

\textbf{Example 3:} \boldorange{Depressed realized prices} due to \boldorange{lack of market access} have forced \boldorange{capital spending cuts}, stalling the \boldorange{growth potential of the company's oil sands assets}.

\textbf{Answer:} \# \boldred{Positive}

\textbf{Explanation:} This is a causal sentence and there exist four events. You need to mark out all four of these events and then pair them up to see if they're related. If so, determine what kind of relationship they belong to. 

\section{Appendix: Out-of-domain Test}
\label{sec:appendix}


\begin{table}[ht]
\centering
\small
\begin{tabular}{lllllll}
\hline
\multirow{2}{*}{\textbf{\begin{tabular}[c]{@{}l@{}}Sec\end{tabular}}} & \multicolumn{2}{c}{\textbf{Consumer}} & \multicolumn{2}{c}{\textbf{Industrial}} & \multicolumn{2}{c}{\textbf{Technology}}  \\
 & {F1} & {Acc} & {F1} & {Acc} & {F1} & {Acc} \\ \hline
\textbf{Con} & \textbf{59.69} & 69.01 & 50.03 & 66.47 & \textit{47.95} & 63.74 \\
\textbf{Ind}  & 50.60 & 67.39 & \textbf{51.14} & 64.61 & 47.49 & 64.09 \\
\textbf{Tec} & 48.65 & 65.87 & 47.70 & 63.56 & \textbf{50.39} & 61.12 \\ \hline
\end{tabular}
\caption{Out-of-domain test results of the BERT-base model for the fine-grained causality classification task.}
\label{table:oodcls}
\end{table}

\begin{table}[ht]
\centering
\small
\begin{tabular}{lllllll}
\hline
\multirow{2}{*}{\textbf{\begin{tabular}[c]{@{}l@{}}Sec\end{tabular}}} & \multicolumn{2}{c}{\textbf{Consumer}} & \multicolumn{2}{c}{\textbf{Industrial}} & \multicolumn{2}{c}{\textbf{Technology}}  \\
 & {F1} & {EM} & {F1} & {EM} & {F1} & {EM} \\ \hline
\textbf{Con} & \textbf{86.26} & 49.54 & 85.75 & 48.37 & 85.20 & 47.26 \\
\textbf{Ind}  & 84.23 & 49.83 & \textbf{86.00} & 50.68 & 84.90 & 47.45 \\
\textbf{Tec} & 86.24 & 47.99 & 86.03 & 47.50 & \textbf{87.09} & 61.12 \\ \hline
\end{tabular}
\caption{Out-of-domain test results of the Span-Large model for the cause-effect extraction task.}
\label{table:oodext}
\end{table}

It has been shown that sector-relevant features from a given domain could become spurious patterns on the other domains, leading to performance decay under distribution shift \cite{ovadia2019can}. We use instances from three sectors with the largest amounts of samples in our dataset for conducting out-of-domain generalization text. These observe in line with recent works revealing that current deep neural models mostly memorize training instances yet struggle to predict on the out-of-distribution data \cite{gururangan2018annotation,kaushik2019learning,srivastava2020robustness}. To evaluate whether methods can generalize on the out-of-distribution data, and to what extent, the results of the out-of-domain test are shown in Table \ref{table:oodcls} and Table \ref{table:oodext}. 

In particular, the model achieves the best performance when the training and test sets are extracted from the articles of the same domain companies. In the out-of-domain test, the model shows varying degrees of performance decay for both tasks. For example, in the fine-grained causality classification task, the model trained with the data from the Consumer Cyclical domain achieves 59.69 F1 Score when testing on the Consumer Cyclical data while decreasing to 47.95 when testing on technology companies. Moreover, in the cause-effect extraction task, the model trained with the data from the Consumer Cyclical domain achieves 86.26 F1 Score when testing on itself while decreasing to 85.20 when testing on Technology. This shows that the domain-relevant patterns learned by the model cannot transfer well between domains.



\end{document}